\documentclass{article}

 \PassOptionsToPackage{numbers, compress}{natbib}
\usepackage[final]{nips_2017}

\usepackage[utf8]{inputenc} % allow utf-8 input
\usepackage[T1]{fontenc}    % use 8-bit T1 fonts
\usepackage{hyperref}       % hyperlinks
\usepackage{url}            % simple URL typesetting
\usepackage{booktabs}       % professional-quality tables
\usepackage{amsfonts}       % blackboard math symbols
\usepackage{nicefrac}       % compact symbols for 1/2, etc.
\usepackage{microtype}      % microtypography
\usepackage{tabularx}
\usepackage{threeparttable}
\usepackage{graphicx}
\usepackage{caption}
\usepackage{lipsum}
\usepackage{float}
\let\OLDthebibliography\thebibliography
\renewcommand\thebibliography[1]{
	\OLDthebibliography{#1}
	\setlength{\parskip}{0pt}
	\setlength{\itemsep}{0pt plus 0.3ex}
}
\title{Determinants of Mobile Money Adoption in Pakistan}

\author{
	Muhammad Raza Khan\\
	       University of California, Berkeley\\
	       mraza@berkeley.edu
	\And
	Joshua Blumenstock\\
	       University of California, Berkeley\\
	       jblumenstock@berkeley.edu
}

\begin{document}
	\maketitle
	\begin{abstract}
%\section{}
In this work, we analyze the problem of adoption of mobile money in Pakistan by using the call detail records of a major telecom company as our input. Our results highlight the fact that different sections of the society have different patterns of adoption of digital financial services but user mobility related features are the most important one when it comes to adopting and using mobile money services. 
\end{abstract}

\section{Introduction}
\label{sec:intro}

%\begin{figure}[tb]%
%	\centering
%	\includegraphics[width=3.3in]{./figs/world2.png}%
%	\caption{Worldwide access to formal financial services, constructed using data from the Global Financial Inclusion Database \cite{demirgucc2015global_inclusion}. Study locations are identified by pins.}
%	\label{fig:worldmap}%
%\end{figure}
%\begin{table*}[!ht]
%	\centering
%	%	\begin{minipage}[t]{1.8in}
%	%	\includegraphics[height=2in]{./figs/ghana.png}
%	%	\end{minipage}
%	%	\quad 
%	\begin{minipage}[b]{0.42\textwidth}
%		\captionsetup{justification=centering}
%		\includegraphics[height=1.8in]{./figs/pakistan.png}
%		\captionof{figure}{Geographic distribution of registered mobile money users in Pakistan. \label{fig:maps}}
%	\end{minipage}
%	\begin{minipage}[b]{0.42\textwidth}
%		\centering
%		\input{tabs/summarystats.tex}
%		
%	\end{minipage}
%	
%\end{table*}

%\input{introduction.tex}
Wider penetration of the mobile phone services in the developing countries is providing alternate ways to provide financial services through mobile phones (``Mobile money services'') in these countries. Considering the lower cost of additional infrastructure government, mobile operators and development agencies all around the world have been trying to promote mobile money services. However, only a few countries have seen widespread adoption of mobile money. Most of the moible money services around the world have been unable to achieve a critical user mass. \cite{scharwatt_state_2014}.
An important question thus revolves around understanding what drives customers to adopt and use mobile money. This paper is focused on the following question: \textit{How do patterns of adoption of mobile money vary across different demographics (gender, urban/rural areas; rich/ poor areas)?} We try to answer this question by applying machine learning algorithms over social networks extracted from mobile money transactions.

%
%\begin{figure}[htb]
%	\centering
%	\begin{minipage}[c]{0.38\textwidth}
%		\centering
%		\input{tabs/summarystats.tex}
%	\end{minipage}
%	\begin{minipage}[c]{0.58\textwidth}
%		\includegraphics[width=\textwidth]{./figs/pakistan.png}
%	\end{minipage}
%	\caption{SOR;\@\vnus{} version, compiled by \rotan{}.}\label{fig:sor-v}
%\end{figure}

%%%%%%%%%%%%%%%%%%%%%%%%%%%%%%%%
% RELATED WORK
%%%%%%%%%%%%%%%%%%%%%%%%%%%%%%%%
\section{Related Work}
\label{sec:related}

%%%%%%%%%%% TABLE %%%%%%%%%%%%%
%\input{tabs/summarystats.tex}
%%%%%%%%%%% TABLE %%%%%%%%%%%%%

%\input{related.tex}
The work on the adoption of digital financial services has been dominated by macroeconomic work related to regulatory issues around interoperability and the logistics of mobile money agents \cite{dermish_branchless_2011,donovan_mobile_2012} and a very few research papers have explored the adoption of mobile money from a quantitative perspective. However, mining of insights from mobile communication meta-data has been a popular area of research and some examples of work in this area include predicting the socioeconomic status \cite{blumenstock_predicting_2015}, gender \cite{frias2010gender}of  mobile phone subscribers, customer churn behavior \cite{khan2015behavioral} and analysis of gender disparities using social networks extracted from mobile communication logs \cite{reed2016gender}.

We have used deterministic finite automata (DFA) based feature engineering over CDR data which is quite similar to the approach used in \cite{khan_dfa_2016} and \cite{blumenstock_predicting_2015}. Lastly, the differences in the usage of technology across men and women, poor \& rich, and the urban \& rural population has been a popular theme of research in the ICTD domain \cite{jackson_race_2008}, \cite{blumenstock2010mobile},etc. 
In comparison to these studies, our work employs a more comprehensive feature engineering approach over a bigger dataset to evaluate the role of the different type of features in the adoption of mobile money for people with different demographic backgrounds.

%%%%%%%%%%%%%%%%%%%%%%%%%%%%%%%%
% DATA
%%%%%%%%%%%%%%%%%%%%%%%%%%%%%%%%

%\label{sec:data}

%\begin{table*}[!ht]
%	\centering
%	%	\begin{minipage}[t]{1.8in}
%	%	\includegraphics[height=2in]{./figs/ghana.png}
%	%	\end{minipage}
%	%	\quad 
%	\begin{minipage}[b]{0.9\textwidth}
%		\centering
%		\input{tabs/sample_statistics_adoption.tex}
%	\end{minipage}
%	%	\begin{minipage}[b]{0.42\textwidth}
%	%		\centering
%	%		\input{tabs/sample_statistics_p2p.tex}
%	%		
%	%	\end{minipage}
%	%	
%\end{table*}

%\input{data.tex}

\section{Context and Methods}
\label{sec:methods}
%\subsection{Feature Engineering }
Our input data consists of anonymized call detail records (CDR) and mobile money transaction information (MMTR) from a major mobile operator of Pakistan. Each row of the CDR typically consists of the tuple containing \texttt{\{callerID, recipientID, date, time, duration, callerLocation, receiverLocation\}}. We categorize each user either as a ``Voice only user",``Registered mobile money user" or ``P2P mobile money user'' as explained below.
\begin{itemize}
	\item \textbf{Voice only users:} The users who do not use mobile money services at all.
	\item \textbf{Registered mobile money users:} The users who have signed up for mobile money.
	\item \textbf{P2P mobile money users:} The users who have either sent or received money to other users.
\end{itemize} 

\begin{figure}[!ht]
	\centering
	%	\begin{minipage}[t]%{0.55\linewidth}
	%	\includegraphics[width=1\textwidth, height=0.2\textheight]{./figs/states.png}
	\includegraphics[width=2.5in]{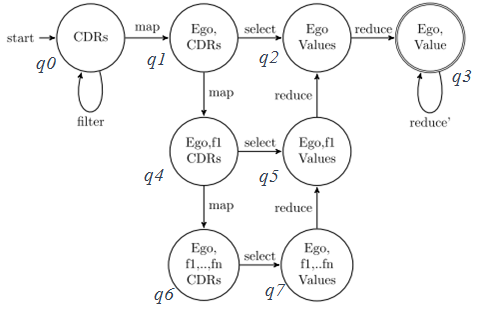}
	\caption{Deterministic Finite Automaton\label{fig:states}\cite{khan_dfa_2016}}	
	%	\end{minipage}
	%	\quad 
	%	\caption{Feature engineering. \label{fig:features}}
\end{figure}

Instead of relying on a few handpicked features, we wanted to use a comprehensive set of features for which we have used the DFA based feature generation algorithm as shown in Figure \ref{fig:states}.
As an example, say we are interested in constructing a feature for each individual $i$ that corresponds to, ``the variance in the average duration of outgoing calls made by $i$ on different days of the week''. This feature can be constructed through the following set of recursive rules: 
(a) filter \textit{outgoing calls},
(b)	filter transactions \textit{initiated by $i$},
(c)	group by \textit{day of week},
(d)	focus on \textit{call duration},
(e)	aggregate by \textit{average} (duration per day of week),
(f) aggregate using \textit{variance} (over average daily durations). DFA also helps in clear categorization of the features. For the purpose of this work, we use the following categories: usage features, mobility features and network (structural properties of the networks) features.
%\subsection{Experimental Design}
We have performed two different type of supervised learning experiments:
\textit{Voice Only Users vs. Registered Mobile Money Users} \&
\textit{Voice Only Users vs. Peer-to-peer Mobile Money Users}

For each of these experiments, we  drew a stratified random balanced sample of adopters/peer-to-peer users and non-adopters for each of the six categories (Males, Females; Urban, Rural; and Rich, Poor ).  Gradient boosting algorithm \cite{friedman2001greedy} is used to classify users and determine prominent features.

\begin{figure*}[!t]
	\centering
	\begin{minipage}[t]{0.31\textwidth}
		\includegraphics[width=\linewidth]{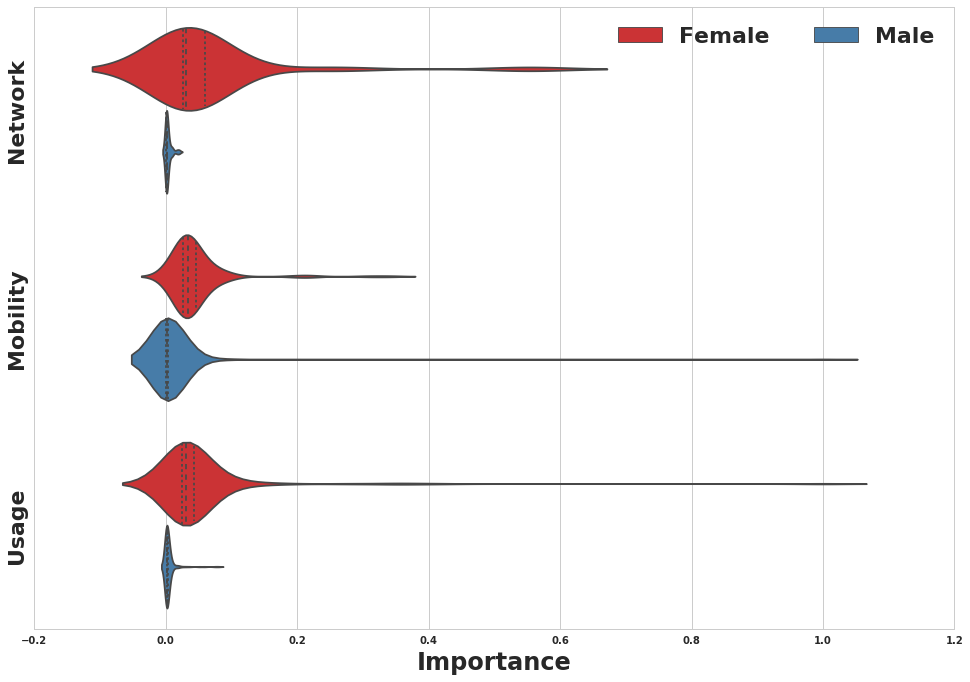}
		\caption{Male vs Female\label{fig:gender_violin}}
	\end{minipage}
	%	\quad 
	\begin{minipage}[t]{0.31\textwidth}
		\includegraphics[width=\linewidth]{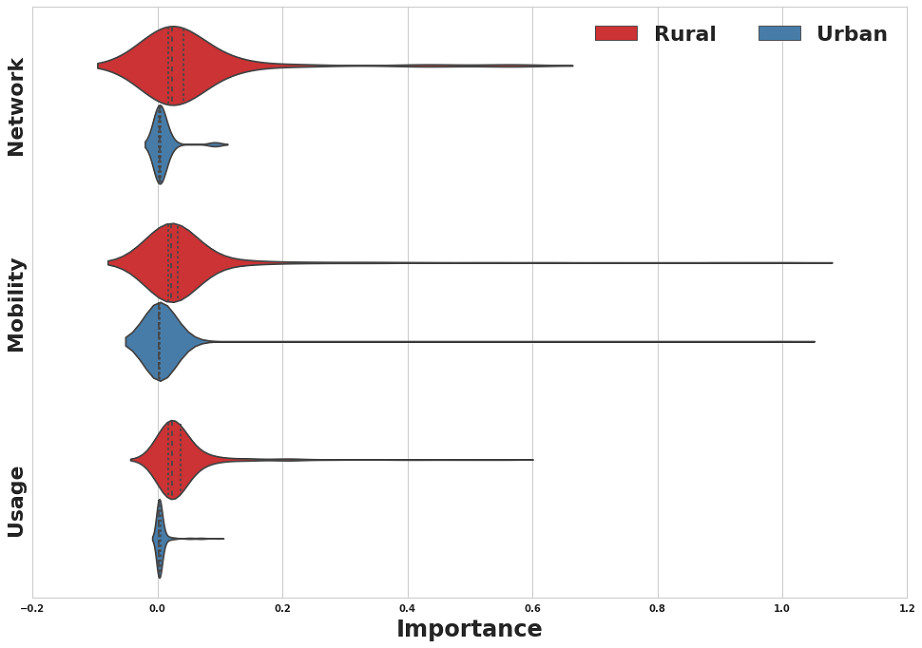}
		\caption{Urban vs Rural Districts\label{fig:urban_violin}}
	\end{minipage}
	%	\quad 
	\begin{minipage}[t]{0.31\textwidth}
		\includegraphics[width=\linewidth]{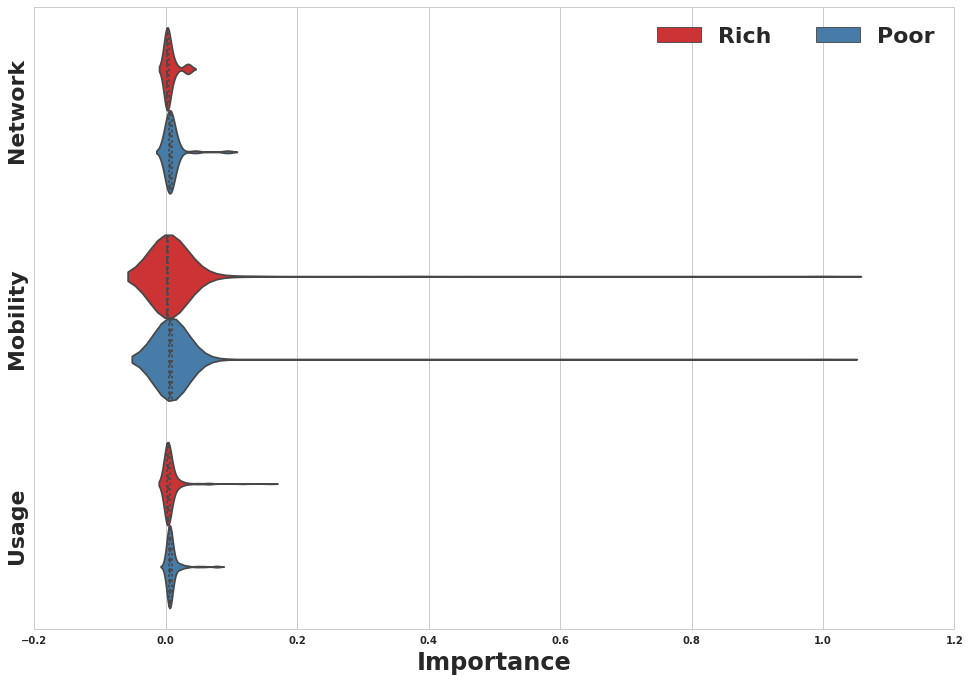}
		\caption{Rich vs Poor Districts \label{fig:mpi_violin}}
	\end{minipage}
	\caption*{Features Importance for Mobile Money Adoption}
\end{figure*}

%%%%%%%%%%%%%%%%%%%%%%%%%%%%%%%%
%% Feature engineering
%%%%%%%%%%%%%%%%%%%%%%%%%%%%%%%%%
%\section{Feature Engineering with Deterministic Finite Automata}
%\label{sec:features}
%
%
%\input{features.tex}

%%%%%%%%%%%%%%%%%%%%%%%%%%%%%%%%
%% METHODS
%%%%%%%%%%%%%%%%%%%%%%%%%%%%%%%%%
%\section{Models and Methods}
%\label{sec:methods}
%\input{methods.tex}

%%%%%%%%%%%%%%%%%%%%%%%%%%%%%%%%
% RESULTS
%%%%%%%%%%%%%%%%%%%%%%%%%%%%%%%%

\begin{figure*}[!t]
	\centering
	\begin{minipage}[t]{0.31\textwidth}
		\includegraphics[width=\linewidth]{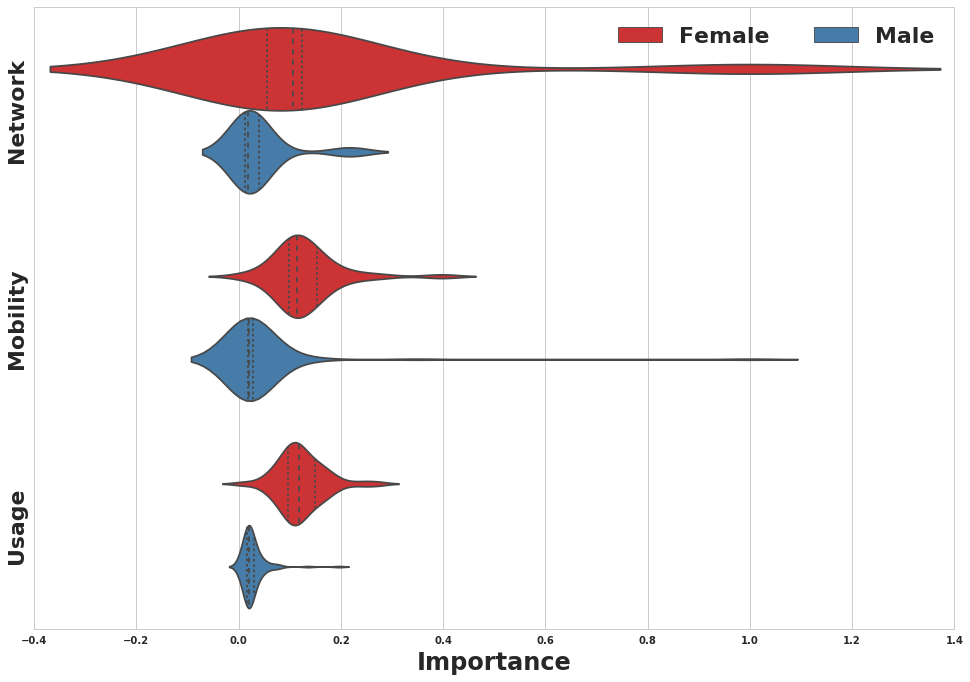}
		\caption{Male vs Female\label{fig:gender_violin_p2p}}
	\end{minipage}
	%	\quad 
	\begin{minipage}[t]{0.31\textwidth}
		\includegraphics[width=\linewidth]{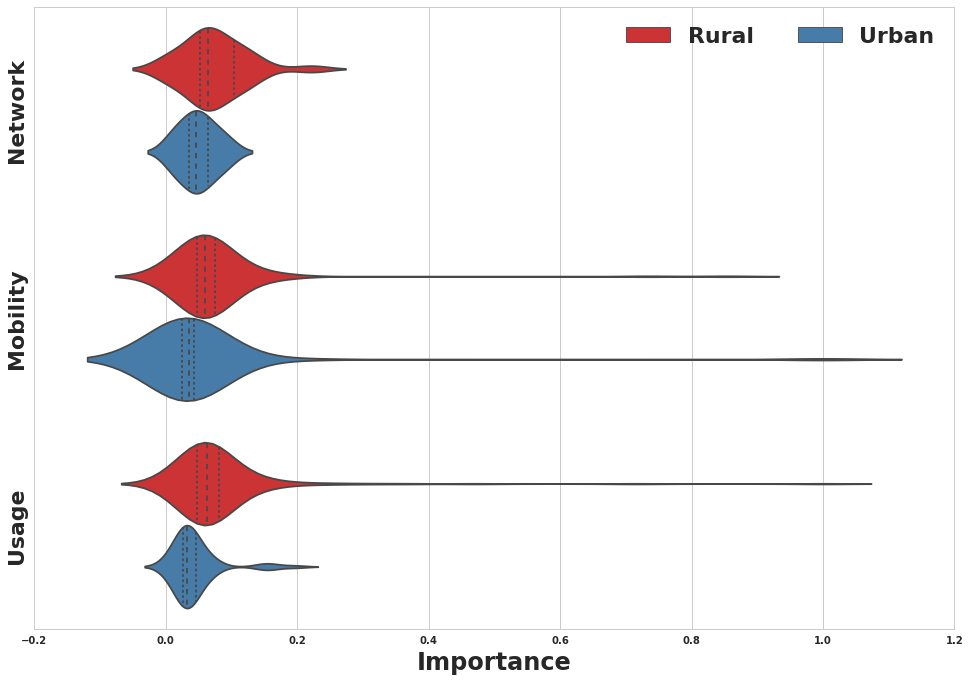}
		\caption{Urban vs Rural Districts \label{fig:urban_violin_p2p}}
	\end{minipage}
	%	\quad 
	\begin{minipage}[t]{0.31\textwidth}
		\includegraphics[width=\linewidth]{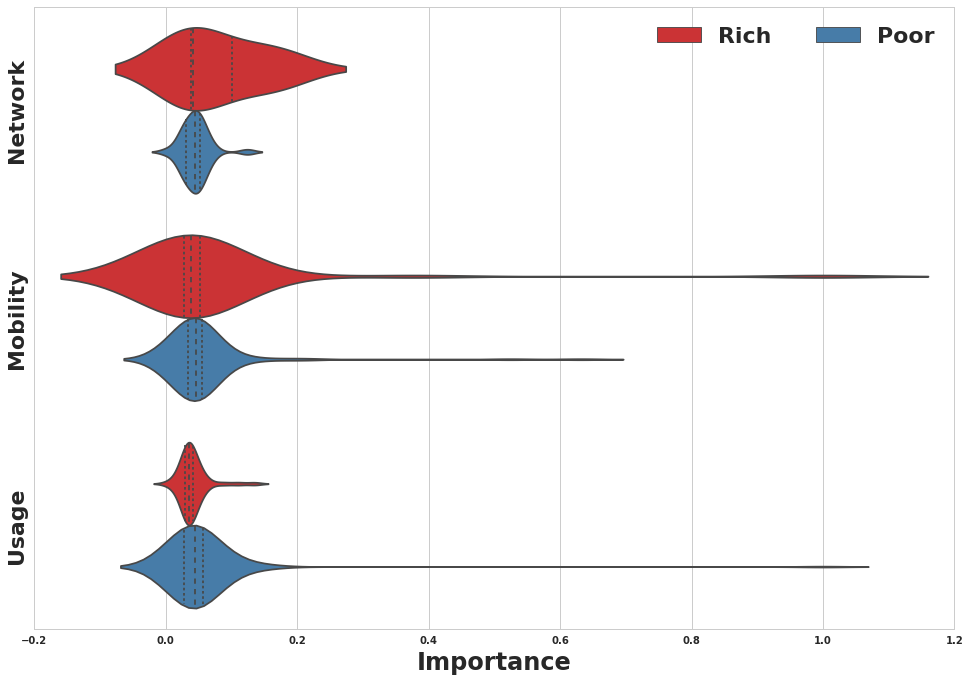}
		\caption{Rich vs Poor Districts \label{fig:mpi_violin_p2p}}
	\end{minipage}
	\caption*{Features Importance for Mobile Money - Peer to Peer Usage}
\end{figure*}

\section{Results}
\label{sec:results}

%\begin{figure}[tb]%
%	\centering
%	\includegraphics[width=3.3in]{./figs/UserCounts.png}%
%	\caption{Distribution of user types by country}%
%	\label{fig:counts}%
%\end{figure}
%
%
%\begin{figure}[tb]%
%		\centering
%		\includegraphics[width=3.3in]{./figs/ghana_calls.png}%
%		\caption{Distribution of calls per subscriber, Ghana}%
%		\label{fig:ghana_calls}%
%\end{figure}

%\input{results.tex}

%\subsection{Determinants of Mobile Money Adoption}
%\subsubsection{Voice only users vs. Registered mobile money users}
The distribution of conditional feature importance values for all the features generated through the DFA is shown in the Figures \ref{fig:gender_violin}, \ref{fig:urban_violin}, \ref{fig:mpi_violin}.  
First thing to note in the Figures \ref{fig:gender_violin}, \ref{fig:urban_violin}, \ref{fig:mpi_violin} is that different category of features have different aggregate feature importance for different cases. Both network and usage related features are a more likely indicator of adoption for females while mobility related features have more predictive power for male users. Usage related features are the most important category for the female users while the mobility related features are the most important one for male users.

Similarly, mobility is the most important category of features for both urban and rural districts. However, compared to rural districts, the network, and usage related features of the users in the urban districts are not that important. The importance of the network related features for the rural users indicates that the awareness of the mobile money services and endorsement effects from other mobile money users play an important role in the adoption of mobile money services. The trends for the rich vs. poor districts are quite similar to the urban and rural districts.

%\subsubsection{Voice only users vs Peer-to-peer mobile money users}
The features importance for each of the experiments for voice only users vs peer to peer mobile money users is shown in the figures \ref{fig:gender_violin_p2p}, \ref{fig:urban_violin_p2p} and \ref{fig:mpi_violin_p2p}.
Usage related features are the top category of peer-to-peer transactions for the female users, while the mobility related features are the top determinant for the male users as shown in the Figure \ref{fig:gender_violin_p2p}. Mobility related features are still the top features for the urban and rural districts (\ref{fig:urban_violin_p2p}) while in comparison to the Figure \ref{fig:urban_violin_p2p} the importance of network related features is lower while the importance of the usage features is higher for the rural district's users. 
In comparison to the Figure \ref{fig:mpi_violin}, Figure \ref{fig:mpi_violin_p2p} shows that the mobility related features are the most important one for the users in the rich districts while the usage related features are the most important determinant for peer to peer transactions in the poor districts.
\begin{figure*}[ht]
	\centering
	\begin{minipage}[t]{0.43\textwidth}
		\includegraphics[width=\linewidth]{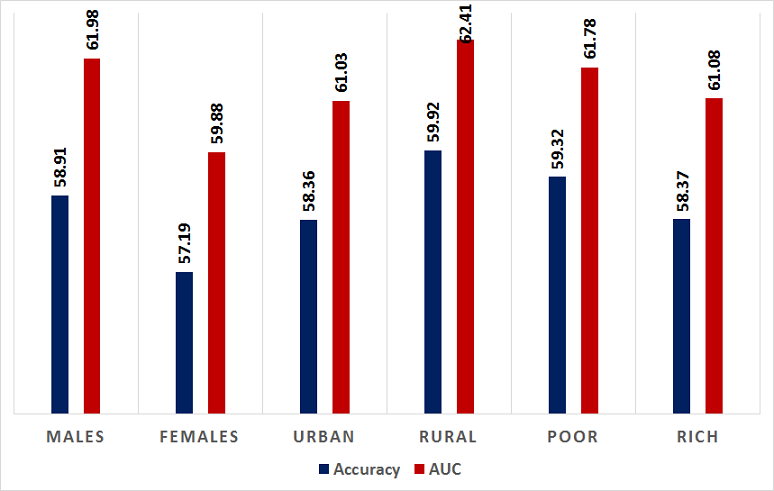}
		\caption{Mobile money adoption accuracy\label{fig:accuracy_adoption}}
	\end{minipage}
	%    \quad 
	\begin{minipage}[t]{0.43\textwidth}
		\includegraphics[width=\linewidth]{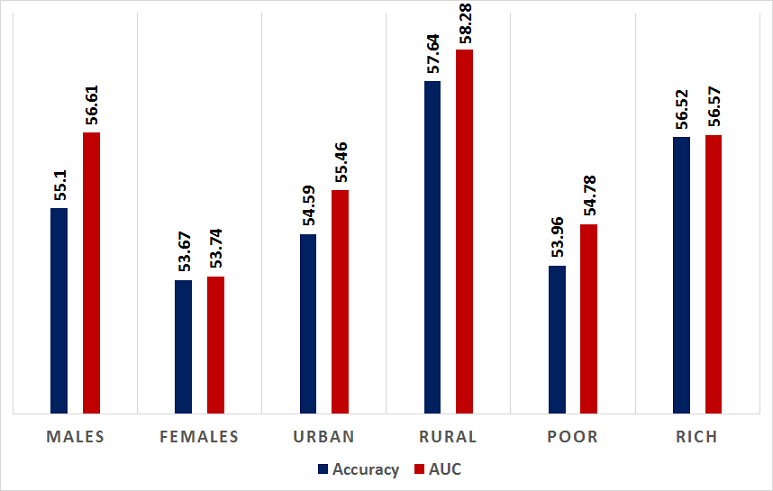}
		\caption{Peer-to-peer usage accuracy\label{fig:accuracy_p2p}}
	\end{minipage}
	
\end{figure*}
Cross-validated accuracy for predicting the adoption or the peer to peer usage of mobile money obtained through gradient boosting is shown in Figure \ref{fig:accuracy_adoption}.

%%%%%%%%%%% TABLE %%%%%%%%%%%%%
%\input{tabs_top_features.tex}

\begin{table*}[!t]\centering
	
	\footnotesize
	%\begin{minipage}[b]{0.32\hsize}\centering
	\begin{tabular}{llc}
		\addlinespace
		\toprule
		\textbf{Index} & \textbf{Feature} & \textbf{Category}  \\
		\midrule
		\multicolumn{2}{l}{\textit{Panel A: Gender: Female}}\\
		1 & Variance in the number of active days with incoming SMS & Usage \\
		2 & Rank Percentile of the Size of the incoming SMS contact network & Network  \\
		% 3 & Rank Percentile of the active days with incoming SMS & Usage \\
		
		\midrule
		\multicolumn{2}{l}{\textit{Panel A: Gender: Male}}\\
		1 & Number of unique contact locations on weekdays & Mobility \\
		2 & Variance in the contact locations per day & Mobility  \\
		% 3 & Number of unique contact locations on weekend & Mobility \\
		\midrule
		\multicolumn{2}{l}{\textit{Panel C: Rural}}\\
		1 & Number of unique contact locations on weekend & Mobility \\
		2& Percentile of the network size on weekend sms  & Network \\
		%3 & Variance in the number of sms per active day &Usage \\
		\midrule
		\multicolumn{2}{l}{\textit{Panel D: Urban}}\\
		1 & Number of unique contact locations on weekdays & Mobility \\
		2& Variance in the duration of the weekend calls to each contact locations  & Usage \\
		%3 & Size of the outgoing calls network on the weekdays &Network \\
		\midrule
		\multicolumn{2}{l}{\textit{Panel E: Poor}}\\
		1 & Number of unique contact locations on week days & Mobility \\
		2& Size of the network per contact locations per day  & Network \\
		%3 & Variance in the number of contact locations per active day &Mobility \\
		\midrule
		\multicolumn{2}{l}{\textit{Panel F: Rich}}\\
		1 & Number of unique contact locations on the weekdays & Mobility \\
		2& Number of unique contact locations on the weekend  & Mobility \\
		%3 & Number of SMS on the weekdays &Usage \\
		\bottomrule
	\end{tabular}
	\begin{tablenotes}[normal]
		\scriptsize
		\item \emph{Notes}: Rank percentile for a feature indicate the percentile of the feature value for the subscriber as compared to all the contacts. 
	\end{tablenotes}
	\caption{Top features for mobile money adoption \label{tab:top_features}}
	%\end{threeparttable}
	
\end{table*}

%%%%%%%%%%% TABLE %%%%%%%%%%%%%

%%%%%%%%%%%%%%%%%%%%%%%%%%%%%%%%
% DISCUSSION
%%%%%%%%%%%%%%%%%%%%%%%%%%%%%%%%
%\section{Discussion}
%\label{sec:discussion}
%
%\input{discussion.tex}

%%%%%%%%%%%%%%%%%%%%%%%%%%%%%%%%
% CONCLUSION
%%%%%%%%%%%%%%%%%%%%%%%%%%%%%%%%
\section{Conclusion}
\label{sec:conclusion}
 Most interesting finding of this work are the differences in performance of the different type of features for different sections of the society. For example, the usage category was the top one for the female users but not for the male users. This indicates that the males who may be working remotely or have to commute to work are more likely to use the mobile money, whereas the women who are more active or have higher technology literacy are more likely to user mobile money. Similarly, mobility of the users is the prime determinant when it comes to the adoption of mobile money in both urban \& rural and rich \& poor districts. However, the people in rural districts with a larger network are more likely to use mobile money.
These results can help the marketing and developing agencies to promote the adoption of mobile money in the developing countries.
% highlight working women as the future work.

%\section*{References}

\bibliographystyle{abbrv}
\bibliography{mm-kdd-references}

\end{document}